# A Proportional-Integral Controller-Incorporated SGD Algorithm for High Efficient Latent Factor Analysis

Jinli Li, Shiyu Long, Minglian Han

*Abstract*—In industrial big data scenarios, high-dimensional sparse matrices (HDI) are widely used to characterize high-order interaction relationships among massive nodes. The stochastic gradient descent-based latent factor analysis (SGD-LFA) method can effectively extract deep feature information embedded in HDI matrices. However, existing SGD-LFA methods exhibit significant limitations: their parameter update process relies solely on the instantaneous gradient information of current samples, failing to incorporate accumulated experiential knowledge from historical iterations or account for intrinsic correlations between samples, resulting in slow convergence speed and suboptimal generalization performance. Thus, this paper proposes a PILF model by developing a PI-accelerated SGD algorithm by integrating correlated instances and refining learning errors through proportional-integral (PI) control mechanism that current and historical information; Comparative experiments demonstrate the superior representation capability of the PILF model on HDI matrices.

*Index Terms*—High-Dimensional and Incomplete Matrix, Stochastic Gradient Descent, Proportional-Integral, Latent Factor Analysis.

## I. Introduction

DATA serves as the foundation for industrial application development. Continuous interactions occur among these nodes, generating substantial information flows [1-6]. Although interactive behaviors are inherently continuous, the information generated remains relatively sparse since each node can only engage with a limited number of peers. Consequently, the High-Dimensional and Incomplete (HDI) matrices are common [7-10].

Although the available data in HDI matrices is known to be relatively sparse, it effectively captures the essential characteristics of each node [11-13, 65-72]. Moreover, extracting distinctive node features from known data within specific HDI matrices can significantly accelerate the development of related applications [51-60]. Among existing methodologies, Latent Factor Analysis (LFA) has emerged as the predominant strategy for HDI information extraction [14-18, 73-76].

Stochastic Gradient Descent (SGD) has become the predominant optimization approach in LFA model implementations [19-22]. However, a critical limitation exists in conventional SGD-LFA frameworks: each parameter update relies exclusively on current-batch learning errors while disregarding historical gradient information, consequently impairing convergence efficiency in industrial-scale applications [17-25, 61-65].

Furthermore, proportional-integral (PI) controllers are employed in over 90% of industrial control scenarios. By systematically incorporating both current system errors and accumulated historical deviations, PI controllers establish comprehensive feedback loops between desired and actual states, thereby enabling precise attainment of control objectives. Inspired by this robust control mechanism, our research innovatively adapts the PI principle to enhance the SGD-LFA framework.

## II. Preliminaries

### A. Problem Formulation

***Definition* 1 (An HDI matrix):** Let two disjoint node sets be denoted as $M$ and $N$ respectively. The entity $r_{m,n}$ represents the inclination degree of node $m \in M$ towards node $n \in N$. The known entity set and unknown entity set are denoted by $\Lambda$ and $\Gamma$ respectively. If $|\Lambda| \ll |\Gamma|$, then $R$ is referred to as an HDI matrix [26-29].

***Definition* 2 (An LFA Model):** The LFA model constructs a low-rank estimation $\hat{R}=XY^T$ to obtain the desired LF matrices $X \in \mathbf{R}^{|M| \times f}$ and $Y \in \mathbf{R}^{|N| \times f}$, which are built based on $\Lambda$. Here, $f \ll \min\{|M|, |N|\}$ denotes the dimensionality of $X$ and $Y$ [30-39]. The cost function of the LFA model is expressed by the Euclidean distance formula shown as below:

$$\begin{aligned}\varepsilon(X,Y) &\triangleq \left( \left(R-\hat{R}\right)^2 + \lambda\left(\|X\|_F^2 + \|Y\|_F^2\right) \right) \\ &= \sum_{r_{m,n} \in \Lambda} \left( \left(r_{m,n}-\hat{r}_{m,n}\right)^2 + \lambda\|x_m\|_2^2 + \lambda\|y_n\|_2^2 \right);\end{aligned} \qquad (1)$$

### B. An SGD Algorithm LFA

Based on previous studies [40-44, 76-80], The SGD's mathematical form is given as Eq.(2):

$$\arg\min_{X,Y} \varepsilon \overset{SGD}{\Rightarrow} \forall r_{m,n} \in \Lambda : x_m \leftarrow x_m - \eta \frac{\partial \varepsilon_{m,n}}{\partial x_m}, y_n \leftarrow y_n - \eta \frac{\partial \varepsilon_{m,n}}{\partial y_n} \qquad (2)$$

Moreover, the complete SGD implementation for the LFA model can be formally expressed as:

During the $t$-th training iteration: $\arg\min\limits_{X,Y} \varepsilon \overset{SGD}{\Rightarrow}$

$$\forall r_{m,n} \in \Lambda : \begin{cases} x_m \leftarrow x_m + \eta \cdot \left(e_{m,n}^{(t)} \cdot y_n - \lambda \cdot x_m\right), \\ y_n \leftarrow y_n + \eta \cdot \left(e_{m,n}^{(t)} \cdot x_m - \lambda \cdot y_n\right), \end{cases} \quad (5)$$

## III. METHODS

### A. The ADRC-based Instant Learning Error Refinement

The SGD-optimized LFA model iteratively refines the LF matrices using individual entries $r_{m,n} \in \Lambda$. As previously defined, $r_{m,n}$ quantifies the preference strength of node $m \in M$ toward node $n \in N$. Since node $m$ typically interacts with numerous nodes in $N$, these interactions collectively capture $m$'s behavioral pattern [81-85]. Therefore, the related nodes of $m$ is given as $r_m$, it contains both observed entries $r_{m,*}$ (known interactions) and missing entries (unknown interactions). Critically, the known entries $r_{m,*}$ are interlinked through $m$'s historical behavior, forming a cohesive representation of $m$'s preferences in the HDI matrix. This justifies treating the known entries of $r_m$ as a unified control target for node $m$. The same logic applies to $r_n$ for nodes in $N$.

Following the established theoretical foundation and incorporating the relationships, we present the node-specific refinement of the SGD algorithm with PI control:

$$\begin{cases} \tilde{e}_m^{(t)} = K_P e_{m,n}^{(t)} + K_I \sum_{k=0}^{t} e_{m,*}^{(k)}, \\ \tilde{e}_n^{(t)} = K_P e_{m,n}^{(t)} + K_I \sum_{k=0}^{t} e_{*,n}^{(k)}, \end{cases} \quad (4)$$

where $\tilde{e}_m^{(t)}$ and $\tilde{e}_n^{(t)}$ represent the refinement learning error of $e_{m,n}^{(t)}$ corresponding to node $m$ and $n$, $e_{m,*}^{(k)}$ and $e_{*,n}^{(k)}$ are calculated by $r_{m,*}$ and $r_{*,n}$, represent the calculated learning error related to node $m$ and $n$, respectively.

The error refinement mechanism consists of:

a) Proportional term: Computed from immediate learning error, reflecting the current node interaction state. Thus, it should be proportional to the learning error.

b) Integral term: Accumulates historical node performance data to improve convergence and dampen oscillations. Therefore, it should be inversely proportional to the learning error.

Thus, the time nodes in the PI controller are replaced with the training rounds of the formula, and $|\Lambda|$ independent PI controllers are adopted respectively for each known instance data to reconstruct the learning error of the current round. Then the learning error reconstruction mode of the instance data $r_{m,n}$ in the $t$-th round of training is as follow:

$$\forall r_{m,n} \in \Lambda : \begin{cases} x_m \leftarrow x_m + \eta \cdot \left(\tilde{e}_m^{(t)} \cdot y_n - \lambda \cdot x_m\right), \\ y_n \leftarrow y_n + \eta \cdot \left(\tilde{e}_n^{(t)} \cdot x_m - \lambda \cdot y_n\right). \end{cases} \quad (5)$$

Therefore, the PILF is completed.

Meanwhile, the computation complexity of PILF depends on the following two factors:

**a) Initialization:** $T_1 = \Theta\left(\left(|M|+|N|\right) \times f\right)$,

**b) Training:** $T_2 = \Theta\left(|\Lambda| \times f + \left(|M|+|N|\right) \times f\right)$,

Therefore, the total computation complexity is given as below:

$$T = T_1 + T_2 \times t \approx \Theta\left(t \times |\Lambda| \times f\right).$$

Moreover, $S$ is storage complexity:

$$S = \left(|M|+|N|\right) \times f.$$

## IV. EXPERIMENTAL RESULTS AND ANALYSIS

### A. General Settings

**Evaluation Protocol**. To conduct a thorough performance assessment, we employ two established metrics: RMSE and MAE to evaluate accuracy. Computational efficiency is evaluated through recorded training durations. Optimal model performance is characterized by minimized error metrics (RMSE/MAE) and reduced computational time, indicating superior convergence properties [45-50]. The mathematical formulations for these metrics are presented as follows:

$$\begin{cases} \text{RMSE} = \sqrt{\left(\sum_{r_{m,n} \in \Phi} (r_{m,n} - \hat{r}_{m,n})^2\right) \Big/ |\Phi|}, \\ \text{MAE} = \left(\sum_{r_{m,n} \in \Phi} |r_{m,n} - \hat{r}_{m,n}|_{abs}\right) \Big/ |\Phi|, \end{cases}$$

where $\hat{r}_{m,n}$ represents the predicted value pertaining to $r_{m,n}$, $\Phi$ represents the testing dataset, $|\ |_{abs}$ represents the operation of taking the absolute value.

**Experimental Datasets.** The experimental validation was conducted using two industrial-scale high-dimensional incomplete (HDI) datasets, with their key characteristics systematically documented in Table 1.

**Compared models.** The evaluation compares our proposed method against two established models as shown in Table 2.

TABLE I. Experimental dataset details.

| No. | Name | Row | Column | Known Entries | Density |
|---|---|---|---|---|---|
| **D1** | ML10M | 71,567 | 10,681 | 10,000,54 | 0.31% |
| **D3** | Douban | 129,490 | 58,541 | 16,830,839 | 0.22% |

## B. Comparison against State-of-the-art Models

Experimental measurements encompassing prediction errors (RMSE/MAE) and computational overhead are presented in Tables 3-4. Moreover, the training curves on D1 are depicted in figure 1.

*a)* The experimental results reveal M1's outstanding convergence speed. For dataset D1, M1's optimization time (28.3s for RMSE, 27.2s for MAE) is substantially shorter than M2's (134.4s, 139.0s) and M3's (5712.2s for both metrics), respectively.

*b)* The experimental data demonstrate M1's superior prediction accuracy. For dataset D1, M1's error metrics (0.7915 for RMSE, 0.6078 for MAE) are significantly better than M2's (0.7939, 0.6131) and M3's (0.8755, 0.6821), respectively.

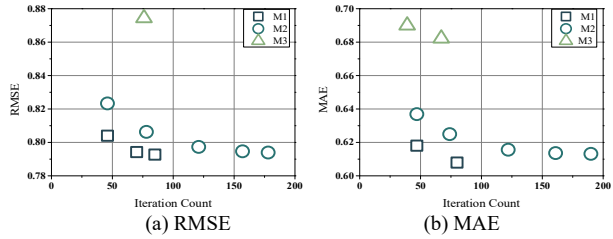

Fig. 1. Convergence curves.

Table 3. Convergence results of RMSE.

|  | **M1** | **M2** | **M3** |
|---|---|---|---|
| **D1-Result** | **0.7915** | 0.7939 | 0.8755 |
|  | **28.3(s)** | 134.4(s) | 5712.2(s) |
| **D2-Result** | **0.7228** | 0.7257 | 0.7625 |
|  | **59.4(s)** | 336.8(s) | 7807.4(s) |

Table 4. Convergence results of MAE.

|  | **M1** | **M2** | **M3** |
|---|---|---|---|
| **D1-Result** | **0.6078** | 0.6131 | 0.6821 |
|  | **27.2(s)** | 139.0(s) | 5712.2(s) |
| **D2-Result** | **0.5605** | 0.5663 | 0.5872 |
|  | **52.4(s)** | 343.2(s) | 7807.4(s) |

TABLE II. Details of compared models.

| Model | Description |
|---|---|
| M1 | The proposed PSLF. |
| M2 | A standard SGD-based LFA model [21]. |
| M3 | A Variational atutoencod-based model [51]. |

TABLE III. Lowest RMSE and their corresponding total time cost (Secs).

| Case | | M1 | M2 | M3 |
|---|---|---|---|---|
| D1 | RMSE: | 0.7910 | 0.7939 | 0.8755 |
| | Time: | 27.3(s) | 134.4(s) | 5712.2(s) |
| D2 | RMSE: | 0.7228 | 0.7257 | 0.7625 |
| | Time: | 59.4(s) | 336.8(s) | 7807.4(s) |

*C. Parameter sensitivity*

This section will conduct parameter sensitivity tests on the gain parameters a and b of PILF. $K_P$ and $K_I$, as the key parameters in the optimization model PILF.

Experimental results are shown in the following figure 2, and the result analysis is as follows:

The parameter KP have a significant promoting effect on the convergence efficiency of the model, but a relatively small promoting effect on the convergence accuracy. With the increase of the KP value, the convergence efficiency of PILF is further improved. However, an excessive KP will affect the convergence accuracy.

The parameter $K_I$ affects the convergence performance. The convergence efficiency of PILF increases with the increase of the $K_I$ value, but the fluctuation range is relatively smaller than the change brought by $K_I$. However, the accuracy fluctuates. This is because the main function of the integration term is to solve the steady-state error to further improve the accuracy, and an overly large $K_I$ value will cause the model to oscillate and exceed the optimal solution.

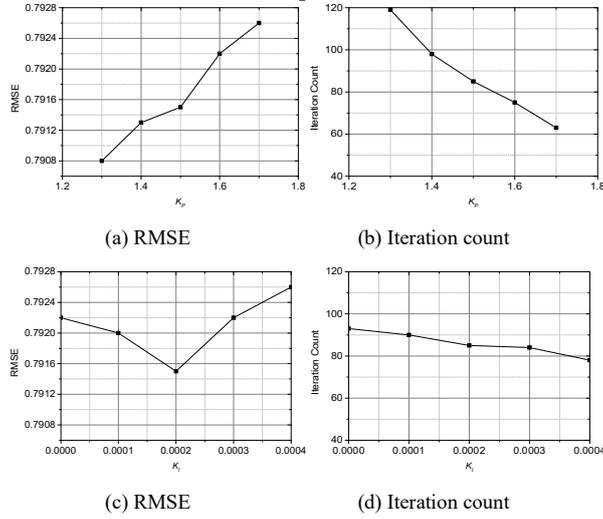

(a) RMSE  (b) Iteration count

(c) RMSE  (d) Iteration count

Fig. 2. Gain parameter sensitivity analysis

## V. CONCLUSIONS

The PILF model introduces PID-controlled node-level optimization for improved latent factor extraction. Validation experiments confirm its advantages in both operational efficiency and predictive performance over baseline approaches. In the future, we will research the intrinsic correlation among samples and further optimize the algorithm from this perspective.